\documentclass{article}
\usepackage{iclr2026_conference,times}
\usepackage[utf8]{inputenc}
\usepackage[T1]{fontenc}
\usepackage{amsmath,amsfonts,amssymb}
\usepackage{graphicx,booktabs,multirow,array,adjustbox}
\usepackage{algorithm,algorithmic}
\usepackage{caption,overpic,xcolor}
\usepackage[most]{tcolorbox}
\usepackage{microtype}
\usepackage{hyperref}
\usepackage{url}
\definecolor{darkred}{RGB}{139,0,0}
\tcbset{colback=gray!10,colframe=black,boxrule=0.4pt,left=6pt,right=6pt,top=4pt,bottom=4pt,sharp corners}
\hypersetup{hidelinks,pdftitle={Scene-R1: Video-Grounded Large Language Models for 3D Scene Reasoning with Lightweight Annotations},pdfauthor={Junjie Wang, Zhihao Yuan, Shuyi Jiang, Chun-Mei Feng, Yaolun Zhang, Shuguang Cui, Zhen Li, Na Zhao}}
\iclrfinalcopy
\title{\raggedright Scene-R1: Video-Grounded Large\\Language Models for 3D Scene Reasoning\\with Lightweight Annotations}
\author{
Junjie Wang$^{1,2,*}$, Zhihao Yuan$^{1,2,*}$, Shuyi Jiang$^{4}$, Chun-Mei Feng$^{3}$,\\
\bfseries Yaolun Zhang$^{1,2}$, Shuguang Cui$^{2,1}$, Zhen Li$^{2,1,\dagger}$, Na Zhao$^{4,\dagger}$\\[6pt]
\normalfont\small $^{1}$ FNii-Shenzhen, CUHKSZ\quad $^{2}$ SSE, CUHKSZ\\
\normalfont\small $^{3}$ University College Dublin\\
\normalfont\small $^{4}$ Singapore University of Technology and Design
}
\begin{document}
\maketitle
\lhead{Scene-R1: Video-Grounded 3D Scene Reasoning}
\begingroup
\renewcommand{\thefootnote}{\fnsymbol{footnote}}
\footnotetext[1]{Equal contribution.}
\footnotetext[2]{Corresponding authors.}
\endgroup
\begin{abstract}
Large language models are increasingly being applied to 3D scene understanding. However, existing 3D-aware LLMs largely operate as black boxes: they produce bounding boxes or textual responses without exposing the reasoning process, and they typically depend on pre-trained 3D detectors for object proposals. We present Scene-R1, a video-grounded framework for 3D scene reasoning that avoids both pre-trained 3D object detectors and dense point-cloud backbones. The framework combines reinforcement-learning-based reasoning with a two-stage grounding pipeline. In the first stage, Scene-R1 performs temporal grounding by identifying the video segments most relevant to an open-ended query. In the second stage, it grounds the target in image space by predicting a 2D bounding box. This prediction is then refined into a precise 3D localization using high-quality candidates from a zero-shot segmentation module, enabling fine-grained geometric reasoning without detector-based proposals. Scene-R1 can also be adapted to 3D visual question answering, generating free-form answers directly from RGB-D video. Our training pipeline uses lightweight 2D supervision projected from sparse scans, avoiding the computational overhead required to train 3D instance segmentation models. Across multiple datasets, Scene-R1 outperforms existing open-vocabulary baselines while also providing interpretable intermediate reasoning steps. These results suggest that reinforcement-learning-based reasoning over RGB-D video can support accurate 3D scene understanding with task-level supervision.
The training, inference, and 3D refinement code is available at \url{https://github.com/819346675/Scene-R1}.
\end{abstract}

\section{Introduction}
\begin{figure}[!t]
\centering
\includegraphics[width=1.0\textwidth]{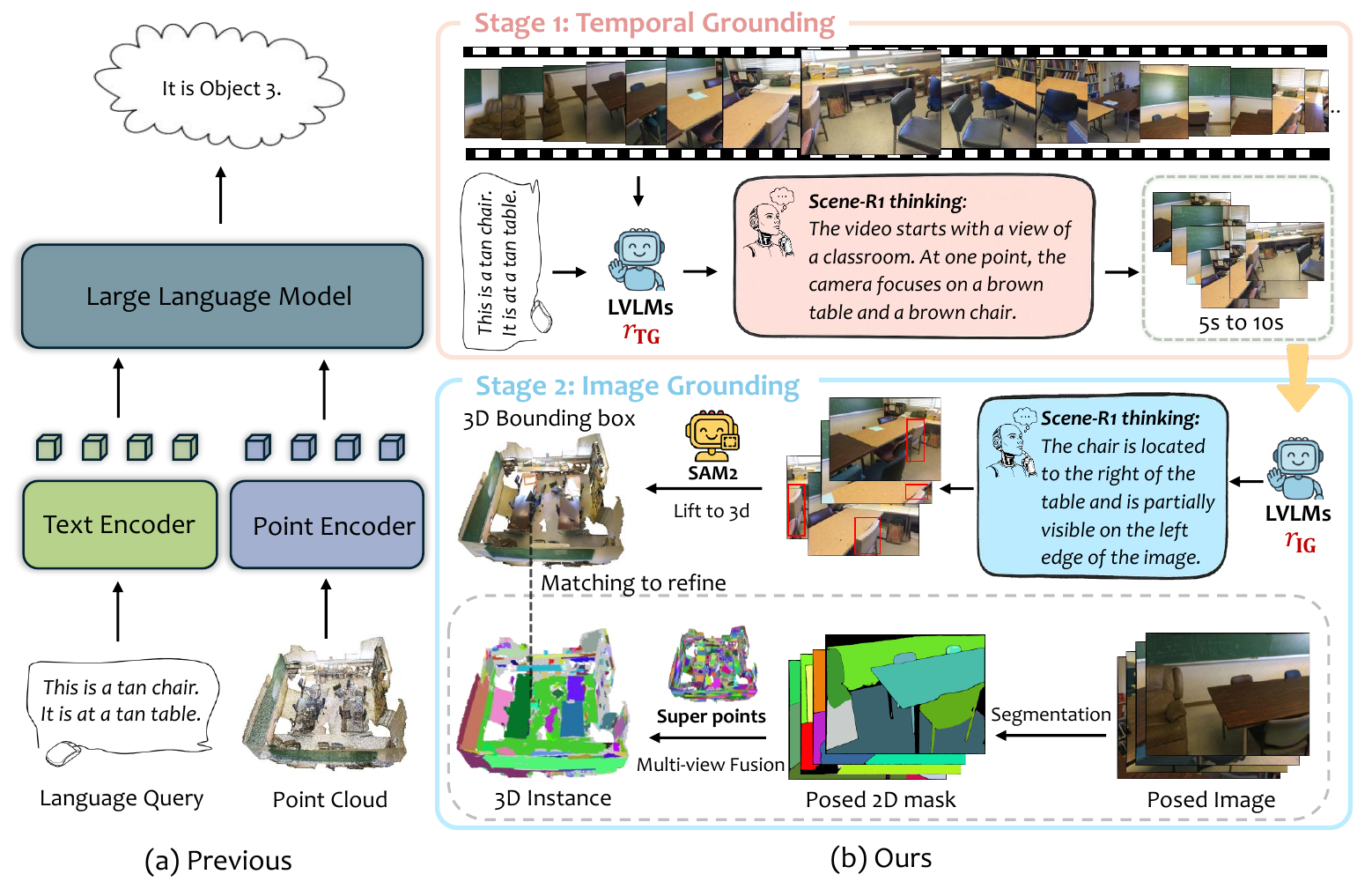} 
\caption{
    \textbf{Overview of Scene-R1 \textit{versus} prior 3D LLMs.}
    \textbf{(a)} Previous 3D LLMs directly output predictions as black boxes without exposing intermediate reasoning, while still depending on pre-trained 3D encoders.
    \textbf{(b)} In contrast, Scene-R1 performs transparent 3D scene reasoning while bypassing the need for dense, point-wise 3D annotations.
}
\label{fig:fig1}
\end{figure}

Large language models (LLMs) have rapidly expanded beyond text to 2D visual perception, with recent applications in embodied AI and robotics ~\citep{vemprala2023chatgpt, brohan2023can}. Extending these models to 3D scene understanding is a natural direction, and recent work has explored 3D visual grounding~\citep{chen2020scanrefer, wang2023distilling, achlioptas2020referit3d}, captioning~\citep{chen2023end, chen2024vote2cap, chen2021scan2cap}, and question answering~\citep{azuma2022scanqa, ma2022sqa3d} on point clouds. Despite this progress, current 3D-aware LLMs (3DLLMs) still face two major limitations, as shown in Fig. \ref{fig:fig1} (a). First, their predictions are difficult to interpret: the models output bounding boxes or short textual answers without exposing the intermediate reasoning process, which complicates debugging and reliability analysis.
Second, most models rely on pre-trained 3D detectors or transformer-based instance segmentation models trained with dense point-level annotations, such as Mask3D~\citep{schult2023mask3d} on ScanNet~\citep{scannet}. Collecting such annotations is costly and difficult to scale to large RGB-D video datasets captured in real-world environments.

Recent advances in reinforcement learning have also improved reasoning capabilities in large language models. Methods such as Group-Relative Policy Optimization (GRPO) and DeepSeek-R1~\citep{guo2025deepseek} encourage models to generate intermediate reasoning steps while improving task performance without requiring manually annotated rationales. Vision-R1~\citep{huang2025vision} extends this paradigm to multimodal settings, showing that reinforcement-learning-based objectives can improve structured reasoning in vision-language models. However, these approaches have primarily focused on text or image understanding. Applying R1-style reinforcement learning to video-based 3D scene understanding remains largely unexplored, particularly in settings that avoid explicit 3D instance detection modules.

We address this gap with Scene-R1, a video-grounded VLM for 3D scene understanding that uses task-level 2D and language supervision while exposing intermediate reasoning steps during inference. As shown in Fig. \ref{fig:fig1} (b), Scene-R1 consists of two grounding stages built on a  shared VLM backbone and optimized with separate RL objectives. In the temporal grounding stage, the model identifies RGB video segments relevant to the input query and produces intermediate reasoning traces. In the image grounding stage, the model predicts a 2D bounding box for the target object in text form. To obtain accurate 3D localization, we further introduce a two-step refinement procedure. The predicted 2D box is first lifted into a coarse 3D bounding box and then refined by selecting the best matching instance from a set of high-fidelity 3D candidates pre-generated by a module inspired by SAI3D~\citep{yin2024sai3d}. This coarse-to-fine process refines the predicted geometry without training a separate 3D instance detector. Our framework can also be adapted to 3D visual question answering by using the similarity between predicted answers and ground-truth responses as the reward signals during RL training. Overall, the results suggest that reinforcement-learning-based reasoning over RGB-D video provides an effective alternative to detector-driven 3D scene understanding pipelines while substantially reducing annotation requirements.

Our main contributions are as follows:
\begin{itemize}
    \item \textbf{RL-based reasoning for 3D scene understanding.} We introduce Scene-R1, a 3D grounding framework trained with R1-style reinforcement learning that produces intermediate reasoning traces during inference, improving the interpretability of 3D-aware VLM predictions.

    \item \textbf{A novel two-stage video-grounded pipeline.} Scene-R1 decomposes 3D grounding into temporal grounding and image grounding stages. The model first selects query-relevant video segments and then localizes target objects in image space, enabling direct reasoning over RGB-D video without requiring a separate 3D detection pipeline.

    \item \textbf{Annotation-efficient 3D localization.} Our method combines RL-based training with coarse-to-fine geometric refinement, allowing the model to learn from 2D supervision while achieving competitive performance on standard 3D indoor scene understanding benchmarks.
\end{itemize}

\section{Related Work}

\subsection{3D Indoor Scene Understanding}
3D indoor scene understanding enables machines, particularly embodied agents, to perceive scene geometry, localize objects, and reason about spatial relationships in indoor environments. These capabilities support several downstream tasks, including: 
1) 3D Visual Grounding~\citep{chen2020scanrefer, wang2023distilling, achlioptas2020referit3d, huang2022multi, jain2022bottom, guo2023viewrefer, chen2022language}, which localizes objects from natural language descriptions; 
2) 3D Dense Captioning~\citep{chen2023end, chen2024vote2cap, chen2021scan2cap}, which generates textual descriptions for multiple objects or regions within a scene;
3) 3D Visual Question Answering~\citep{azuma2022scanqa, ma2022sqa3d},  requires models to answer language queries grounded in 3D geometry and semantics; and
4) Affordance Grounding~\citep{delitzas2024scenefun3d, corsetti2025fun3du}, which predicts potential human-object interactions based on object functionality and scene context.
Early Methods~\citep{ma2022sqa3d,azuma2022scanqa,chen2020scanrefer,achlioptas2020referit3d} typically focused on individual tasks, whereas more recent work has explored unified frameworks~\citep{huang2024embodied, huang2024chat} that support multiple forms of 3D scene understanding within a shared model.

\subsection{Multi-modal Large Language Models}
Multi-modal large language models (MLLMs) extend conventional LLMs by incorporating inputs from multiple modalities, including images~\citep{radford2021learning, rombach2021highresolution, zhao2023bubogpt}, audio~\citep{zhao2023bubogpt}, video~\citep{li2023videochat, wang2024grounded}, and 3D point clouds~\citep{peng2023openscene,fu2024scene, xu2024pointllm, guo2023point}, in addition to text. 
In the 3D domain, LERF~\citep{kerr2023lerf} learns a dense, multi-scale language field within a NeRF representation by rendering CLIP embeddings along training rays. However, the method relies on per-scene optimization, which limits scalability and generalization across environments. 
OpenScene~\citep{peng2023openscene} aligns 3D scene representations with 2D vision-language models through cross-modal distillation, enabling open-vocabulary semantic segmentation and grounding.
Chat-Scene~\citep{huang2024chat, yuan2025empowering} represents scenes using object-centric features and identifiers, allowing language models to operate directly over object-level representations.

Despite recent progress, most existing 3D-aware LLMs still face two main limitations. First, many methods rely on pre-trained 3D detectors or segmentation models, such as Mask3D~\citep{schult2023mask3d}, which require dense point-level annotations that are expensive and difficult to scale. Second, these models typically provide only final predictions, such as bounding boxes or textual answers, without exposing intermediate reasoning steps, making their decisions difficult to interpret. In contrast, our method does not train a dense 3D segmentation backbone or depend on pre-trained 3D instance detectors. We instead introduce a two-stage grounding framework that performs 3D understanding directly from RGB-D video. By relying on task-level 2D and language supervision, our approach reduces annotation cost while also producing interpretable intermediate reasoning traces.

\subsection{Reinforcement Learning}
With growing interest in model interpretability and reasoning-oriented models such as DeepSeek-R1~\citep{guo2025deepseek} and OpenAI-o1~\citep{openai2024o1}, recent work has explored the use of reinforcement learning (RL) to improve reasoning in VLMs. 
R1-V~\citep{chen2025r1v} demonstrates that RL training can improve generalization on image reasoning tasks for large vision-language models. 
Timezero~\citep{wang2025timezero} applies RL to video-language reasoning and improves temporal grounding performance. 
Video-R1~\citep{feng2025video} introduces large-scale video reasoning datasets together with the T-GRPO training algorithm, which encourages models to use temporal information by contrasting ordered and shuffled video frames during training. 
Despite these advances, the use of reinforcement learning for 3D indoor scene understanding remains relatively unexplored. In this work, we introduce a 3D grounding framework trained with R1-style reinforcement learning that produces intermediate reasoning traces during inference, improving the interpretability of 3D-aware vision-language models.

\section{Preliminary}

\subsection{3D Large Language Models}

Contemporary 3D-aware vision-language models typically follow an instance-centric pipeline, as shown in Fig. \ref{fig:fig1} (a). The pipeline begins with a point-cloud object detector, usually an instance segmentation model such as Mask3D \citep{schult2023mask3d}, which decomposes the scene into $n$ object masks $\{\mathcal{P}_i\}_{i=1}^{n}$.
For each object, a 3D encoder (e.g., Uni3D \citep{zhou2023uni3d}) extracts geometric features, while a 2D encoder captures appearance information from multi-view images. Lightweight visual-language projection layers then map these features into the language token space.
The resulting object identifiers, together with the projected 2D and 3D features, are concatenated into a scene-level token sequence that the language model uses to ground queries and generate responses.
Because this pipeline depends on an initial object detection stage trained with dense point-level instance annotations, it inherits substantial annotation cost and is typically limited to a small number of curated datasets. In contrast, Scene-R1 removes the dependence on explicit 3D instance detectors and instead learns from RGB-D video using task-level supervision.

\subsection{DeepSeek-R1 and Group-Relative Policy Optimization}
\label{sec:grpo}

DeepSeek-R1-Zero, presented alongside DeepSeek-R1 \citep{guo2025deepseek}, demonstrates reasoning-oriented reinforcement learning without a preliminary supervised fine-tuning stage. DeepSeek-R1 additionally uses supervised data in its training pipeline.
Its optimization backbone, \emph{Group-Relative Policy Optimization}
(\textbf{GRPO}) \citep{shao2024deepseekmath} departs from actor–critic methods such as PPO \citep{schulman2017proximal} in two
respects.

\subsubsection{Sampling and group-normalized reward}
For every prompt $p$ the current policy $\pi_\theta$ generates a \emph{group} of
$G$ candidate responses $\mathbf{o}=\{o_1,\dots,o_G\}$.
A task-specific scalar reward function $r(\cdot)$ is applied to each response,
yielding $\{r(o_i)\}_{i=1}^{G}$.
GRPO converts raw rewards into \emph{relative} scores.
\begin{equation}
\bar r(o_i)=\frac{r(o_i)-\mu}{\sigma},\quad \sigma>0. 
\end{equation}
\begin{equation}
\mu=\tfrac1G\textstyle\sum_{j=1}^{G} r(o_j),\;
\end{equation}
\begin{equation}
\sigma=\sqrt{\tfrac1G\sum_{j=1}^{G} (r(o_j)-\mu)^2}. 
\end{equation}
The expression above describes non-degenerate groups; a practical implementation must handle zero-variance rewards. The group-wise standardization makes the update
invariant to the reward scale and emphasizes responses that outperform their peers.

\subsubsection{Weighted objective}
Let $\pi_{\theta_\text{old}}$ denote the policy parameters from the previous optimization step.
The group-normalised rewards are combined with an importance-sampling ratio to form
\begin{equation}
R(\mathbf{o})
        =\sum_{i=1}^{G}
          \frac{\pi_\theta(o_i)}{\pi_{\theta_\text{old}}(o_i)}\,
          \bar r(o_i),
\label{eq:grpo_r}
\end{equation}
where $\pi_\theta(o_i)$ is the probability assigned by the current policy to
response $o_i$.
A simplified response-level view of the KL-regularised objective is
\begin{equation}
\max_{\theta}\;
\mathbb{E}_{\mathbf{o}\sim\pi_{\theta_\text{old}}(p)}
\!\Bigl[R(\mathbf{o})-\beta\,D_{\mathrm{KL}}\!\bigl(\pi_\theta\;\|\;\pi_\mathrm{ref}\bigr)\Bigr],
\label{eq:grpo_obj}
\end{equation}
where $\pi_\mathrm{ref}$ is the frozen reference model and
$\beta$ controls the trust-region size.
Crucially, Eq.~\eqref{eq:grpo_obj} obviates the need for a learned critic,
reducing memory overhead and stabilising optimization when reward
computation—e.g.\ IoU over long videos—is costly.
Group normalization reduces sensitivity to a common positive scaling of the total reward; it does not remove the need to specify the relative weights of distinct reward components.

\section{Method}
\label{method}

We address the task of grounding free-form textual queries in RGB-D video and reasoning about the referred 3D regions without training a point-cloud instance detector, as shown in Fig. \ref{fig:fig1} (b).  Formally, each scan has a corresponding video $\mathcal{V}=\{(I_t, D_t, \mathbf{T}_t)\}_{t=1}^{T}$, where $I_t$ and $D_t$ are color and depth images and $\mathbf{T}_t$ the camera pose; $\mathbf{K}$ denotes fixed intrinsics.  Given a query $q$, the model must (i) return the 3D region that $q$ describes, (ii) answer any accompanying question, and (iii) supply a chain-of-thought explanation. Scene-R1 is built on the publicly released Qwen2.5-VL \citep{bai2025qwen2} backbone, whose instruction tuning already encompasses 2D grounding and
general VQA.  We exploit this prior to teach it to understand the 3D world and minimise the amount of
task-specific reinforcement learning (RL).  The same architecture is used in all tasks optimized with GRPO. 

\subsection{Stage 1: Temporal Grounding}

For 3D visual grounding and affordance grounding, we employ a two-stage pipeline.
Given a natural-language description, the model first predicts a continuous time window 
\[
\hat S=[\hat t_s,\;\hat t_e]
\]
which should enclose all frames containing the target object.
Let $\mathrm{fps}$ be the video frame rate.  
The window endpoints are converted to integer frame indices,
\[
\hat f_s=\bigl\lfloor \hat t_s \cdot \mathrm{fps} \bigr\rfloor, \qquad
\hat f_e=\bigl\lceil  \hat t_e \cdot \mathrm{fps} \bigr\rceil,
\]
yielding the predicted frame set
\(
\hat{\mathcal L}= \{\hat f_s,\;\hat f_s+1,\dots,\hat f_e\}.
\)
The ground-truth set of relevant frames is
$\mathcal L^{\star}=\{\ell_1,\dots,\ell_{|\mathcal L^{\star}|}\}\subset\{1,\dots,T\}$.
We measure overlap via the frame-level Intersection-over-Union
\begin{equation}
\operatorname{IoU}_{\text{time}}
  (\hat{\mathcal L},\mathcal L^{\star})
  =\frac{\bigl|\hat{\mathcal L}\cap\mathcal L^{\star}\bigr|}
         {\bigl|\hat{\mathcal L}\cup\mathcal L^{\star}\bigr|}.
\label{eq:iou_time}
\end{equation}
In addition, answers must follow the template
\texttt{<think>\dots</think> <answer> <$t_s$> to <$t_e$></answer>} to first output the chain of thoughts, then the answers.
We therefore define a format reward
\begin{equation}
r_{\text{form}}(o)=
   \begin{cases}
     1,& \text{if the output \(o\) matches the template;}\\[2pt]
     0,& \text{otherwise.}
   \end{cases}
\label{eq:r_form}
\end{equation}
and pass to GRPO the combined reward
\begin{equation}
r_{\text{TG}}
   = \operatorname{IoU}_{\text{time}}\!\bigl(\hat{\mathcal L},\mathcal L^{\star}\bigr)
     \;+\;
     \lambda\,r_{\text{form}}(o),
\qquad
\lambda = 0.1.
\label{eq:r_tg}
\end{equation}
The resulting window \(\hat S\) along with the corresponding RGB-D frames is then forwarded to Stage 2 for image-level object localization.

\subsection{Stage 2: Image Grounding}
\label{sec:image}
Conditioned on the temporal window $\hat S$ obtained in Stage 1, the model now
localizes the target object in every retained frame.
For each frame index $\tau\in\hat S$, the Qwen2.5-VL decoder is prompted to output a JSON line: 
\texttt{\{"bbox\_2d": [x$_1$,\,y$_1$,\,x$_2$,\,y$_2$], "label": <object-class>\}},
where $(x_1,y_1)$ and $(x_2,y_2)$ denote the top-left and bottom-right pixel coordinates of the predicted box
$\hat{\mathbf b}_{\tau} = (x_1,y_1,x_2,y_2).$
Given the ground-truth box $\mathbf b^{\star}_{\tau}$ for frame $\tau$, we compute the spatial Intersection-over-Union
\begin{equation}
\operatorname{IoU}_{\mathrm{box}}
   \bigl(\hat{\mathbf b}_{\tau},\,\mathbf b^{\star}_{\tau}\bigr)
   =\frac{\lvert\,\hat{\mathbf b}_{\tau}\cap\mathbf b^{\star}_{\tau}\rvert}
         {\lvert\,\hat{\mathbf b}_{\tau}\cup\mathbf b^{\star}_{\tau}\rvert}.
\label{eq:iou_box}
\end{equation}
We reuse the \texttt{think/answer} format reward from Stage 1, and introduce an additional JSON-specific reward
\begin{equation}
r_{\text{json}}(o)=
   \begin{cases}
     1,& 
         \text{valid JSON with the schema above};\\[2pt]
     0,& \text{otherwise.}
   \end{cases}
\label{eq:r_json}
\end{equation}
The final reward passed to GRPO is
\begin{equation}
r_{\text{IG}}
   = \operatorname{IoU}_{\mathrm{box}}
        \bigl(\hat{\mathbf b}_{\tau},\mathbf b^{\star}_{\tau}\bigr)
     \;+\;
     \lambda \bigl(r_{\text{json}}(o)\;+\;r_{\text{form}}(o)\bigr).
\label{eq:r_ig}
\end{equation}
This reward encourages spatial overlap and compliance with the output format and JSON schema. The format terms do not directly evaluate the correctness of the generated reasoning.

\subsection{Lifting 2D Predictions to 3D}
\label{sec:lift3d}

To obtain a true 3D localization, we turn each frame-wise box into a dense mask, back-project the masked pixels, and unite the resulting points across the entire
temporal window~\(\hat S\).
For every predicted box \(\hat{\mathbf b}_{\tau}\) at frame \(\tau\), we feed the RGB image and the box coordinates to
SAM2 \citep{ravi2024sam} as a prompt.
SAM2 propagates the cue over neighbouring frames, yielding a binary mask
\(M_{\tau}\!\in\!\{0,1\}^{H\times W}\) where \(M_{\tau}(u,v)=1\) marks object pixels.
Each foreground pixel \((u,v)\) with depth \(D_{\tau}(u,v)\) is converted to a
3D point in the camera coordinate system,
\begin{equation}
\mathbf{x}_{\text{cam}}
  = D_{\tau}(u,v)\;
    \mathbf K^{-1}
    \begin{bmatrix}
      u \\ v \\ 1
    \end{bmatrix},
\label{eq:x_cam}
\end{equation}
where \(\mathbf K\) is the intrinsic matrix.
The point is then lifted into the world coordinates via the camera pose
\(\mathbf T_{\tau}\in\mathbb{R}^{4\times4}\):
\begin{equation}
\mathbf X
  = \mathbf T_{\tau}
    \begin{bmatrix}
      \mathbf{x}_{\text{cam}} \\ 1
    \end{bmatrix}.
\label{eq:x_world}
\end{equation}
Aggregating all such points over \(\tau\in\hat S\) produces the instance-level point cloud
\begin{equation}
\mathbf P
  \;=\;
  \bigcup_{\tau\in\hat S}
    \bigl\{\mathbf X \mid M_{\tau}(u,v)=1\bigr\}.
\label{eq:point_cloud}
\end{equation}
While the tightest axis-aligned bounding box around $\mathbf P$ provides an initial 3D localization, the lifted points can be incomplete or noisy. We refine this localization using scene-level instance candidates pre-generated by SAI3D \citep{yin2024sai3d}, which merges geometric primitives based on multi-view 2D mask consistency. In the accompanying refinement implementation, the lifted target points and each candidate are voxelized in a shared world coordinate system. We select the candidate with the largest voxel intersection-over-union with $\mathbf P$ and return its tight axis-aligned bounding box. If no candidate has positive overlap, the coarse localization is retained. Candidate selection uses the predicted target geometry, without consulting ground-truth instance identities. This candidate-matching step is distinct from the 3D box IoU used to evaluate visual grounding.

\subsection{3D Visual Question Answering}
Scene-R1 can be fine-tuned for 3-D visual question answering (3DVQA) with only minor changes to the reinforcement objective.  
The \texttt{<think>} block provides the chain of thought, while the \texttt{<answer>} tag encloses the final answer \(a\).
Let \(a^{\star}\) denote the ground-truth answer.  
We adopt two complementary metrics from \citet{huang2024embodied}:
\begin{itemize}
\item Exact Match (EM): \(\operatorname{EM}(a,a^{\star})=1\) iff the normalised strings are identical.
\item Refined Exact Match (EM-R): a soft variant that tolerates minor lexical variation.
\end{itemize}
We reuse the \texttt{think/answer} formatting reward \(r_{\text{form}}(o)\) introduced in Stage 1.
The scalar reward supplied to GRPO is
\begin{equation}
r_{\text{QA}}
   = \operatorname{EM}(a,a^{\star})
     + \operatorname{EM\hbox{-}R}(a,a^{\star})
     + \lambda\,r_{\text{form}}(o).
\label{eq:r_qa}
\end{equation}

This one-stage formulation rewards answer agreement and output-format compliance. It does not directly assess explanatory clarity or the faithfulness of the generated reasoning.

\begin{table}[t]
    \centering
    \caption{3DVG results on ScanRefer validation set. The accuracy on the ``unique" subset, ``multiple" subset, and whole validation set is all provided. Following \citep{chen2020scanrefer}, we label the query as ``unique" if it only contains a single object of its class. Otherwise, we label it as ``multiple".}
    \label{tab:tab1}

    \begin{adjustbox}{max width=\linewidth}
\begin{tabular}{lccccccc}
        \toprule 
        & & \multicolumn{2}{c}{ Unique } & \multicolumn{2}{c}{ Multiple } & \multicolumn{2}{c}{ Overall } \\
        \cmidrule(lr){3-4} \cmidrule(lr){5-6} \cmidrule(lr){7-8}
        Methods & Agent & Acc@0.25 & Acc@0.5 & Acc@0.25 & Acc@0.5 & Acc@0.25 & Acc@0.5 \\
        \midrule

        \multicolumn{8}{l}{\textbf{Fully Supervised Methods}} \\
        \addlinespace
        ScanRefer \citep{chen2020scanrefer} & -- & 65.0 & 43.3 & 30.6 & 19.8 & 37.3 & 24.3 \\
        TGNN \citep{huang2021text} & -- & 64.5 & 53.0 & 27.0 & 21.9 & 34.3 & 29.7 \\
        InstanceRefer \citep{yuan2021instancerefer} & -- & 77.5 & 66.8 & 31.3 & 24.8 & 40.2 & 32.9 \\
        3DVG-Transformer \citep{zhao20213dvg} & -- & 81.9 & 60.6 & 39.3 & 28.4 & 47.6 & 34.7 \\
        BUTD-DETR \citep{jain2022bottom} & -- & 84.2 & 66.3 & 46.6 & 35.1 & 52.2 & 39.8\\
        ConcreteNet \citep{unal2024four} & -- & 86.3 & 82.1 & 42.4 & 38.4 & 50.6 & 46.5 \\
        
        \midrule

        \multicolumn{8}{l}{\textbf{LLMs w/ 3D Inst. Supervision Dependency}} \\
        \addlinespace
        ZSVG3D \citep{yuan2024visual} & GPT-4 turbo & 63.8 & 58.4 & 27.7 & 24.6 & 36.4 & 32.7 \\
        SeeGround \citep{li2025seeground} & Qwen2-VL-72b & 75.7 & 68.9 & 34.0 & 30.0 & 44.1 & 39.4 \\
        LLaVA-3D \citep{zhu2024llava} & LLaVA-Video-7B & -- & -- & -- & -- & 54.1 & 42.4 \\
        Video-3D LLM \citep{zheng2025video} & LLaVA-Video-7B & -- & -- & -- & -- & 58.1 & 51.7 \\
        3D-LLaVA \citep{deng20253d} & LLaVA-1.5-7B & -- & -- & -- & -- & 51.2 & 40.6 \\
        \midrule
        \multicolumn{8}{l}{\textbf{Label-Efficient or Zero-Shot Methods}} \\
        \addlinespace
        LERF \citep{kerr2023lerf} & CLIP & - & - & -  & - & 4.8 & 0.9 \\
        OpenScene \citep{peng2023openscene} & CLIP & 20.1 & 13.1 & 11.1 & 4.4 & 13.2 & 6.5 \\
        VLM-Grounder(Zero-Shot) \citep{xu2025vlm} & GPT-4o & \textbf{66.0} & 29.8 & 48.3 & 33.5 & 51.6 & 32.8 \\
        \textbf{Ours (Scene-R1)} & Qwen2.5-VL-7B & 64.1 & \textbf{49.1} & \textbf{50.4} & \textbf{39.4} & \textbf{53.1} & \textbf{41.2} \\
        \bottomrule
    \end{tabular}
\end{adjustbox}
\end{table}

\begin{table}[t]
\centering
\small
\caption{Evaluation of 3D Question Answering on SQA3D~\citep{ma2022sqa3d}. We demonstrate accuracy on six types of questions based on their prefixes.}
\label{tab:tab2}
{
\begin{adjustbox}{max width=\linewidth}
\begin{tabular}{lccccccccc}
\toprule
Methods & Supervision & What & Is & How & Can & Which & Others & Avg. (EM-1) & EM-R1 \\
\midrule
ClipBERT~\citep{ma2022sqa3d} & fully  & 30.2 & 60.1 & 38.7 & 63.3 & 42.5 & 42.7 & 43.3 & -- \\
SQA3D~\citep{ma2022sqa3d} & fully  & 31.6 & 63.8 & \textbf{46.0} & 69.5 & 43.9 & 45.3 & 46.6 & -- \\
3D-VisTA~\citep{zhu20233d} & fully  & 34.8 & 63.3 & 45.4 & \textbf{69.8} & 47.2 & 48.1 & 48.5 & -- \\

PQ3D~\citep{zhu2024unifying} & fully  & 37.1 & 61.3 & 44.5 & 60.9 & 47.0 & 45.1 & 47.1 & -- \\
LEO~\citep{huang2024embodied} & fully  & 39.0 & \textbf{63.9} & 44.9 & 66.2 & 47.7 & 51.1 & \textbf{50.0} & 52.4 \\
\hline
Ours & - & \textbf{41.9} & 62.1 & 40.1 & 64.1 & \textbf{48.3} & \textbf{51.2} &  49.7 & \textbf{53.2} \\
\bottomrule
\end{tabular}
\end{adjustbox}
}
\end{table}

\section{Experiments}
\label{sec:experiments}

This section assesses Scene-R1 on three core 3D scene understanding tasks: visual grounding, affordance grounding, and visual question answering.
We first describe implementation details and evaluation
protocols (Sec. \ref{sec:impl}), then present quantitative (Sec. \ref{sec:main_results}) and qualitative (Sec. \ref{sec:qualitative}) results, followed by an ablation study that
analyses design choices (Sec. \ref{sec:abl}).

\subsection{Implementation Details}
\label{sec:impl}

\subsubsection{Datasets}
ScanRefer \citep{chen2020scanrefer} extends ScanNet \citep{scannet} RGB-D reconstructions with free-form referring expressions that localize objects directly in 3D space.  
The corpus contains 51,583 descriptions referring to 11,046 target objects across 800 indoor scenes \citep{scannet}.  
SceneFun3D \citep{delitzas2024scenefun3d} targets fine-grained functionality understanding in real-world indoor scans captured with a high-resolution Faro laser scanner \citep{baruch2021arkitscenes}.  
VSI-Bench \citep{yang2025thinking} is a video-based benchmark designed to evaluate the visual-spatial intelligence of Multimodal Large Language Models (MLLMs). It includes over 5,000 question-answer pairs derived from 288 real-world indoor videos sourced from ScanNet, ScanNet++, and ARKitScenes. The benchmark features eight distinct tasks, such as object counting, route planning, and distance estimation, which test an MLLM's ability to see, remember, and recall spatial information from sequential visual input.

\subsubsection{Training schedule}
We initialise all experiments from the \texttt{Qwen2.5-VL-7B} checkpoint \citep{bai2025qwen2}.  
The model ingests a sequence of interleaved visual and textual tokens.  
RGB frames are sampled at 2 fps; ScanNet clips are resized to $640{\times}480$ while ARKitScenes clips retain their native resolution.  
To bound the computational footprint across heterogeneous inputs we employ the \emph{smart-resize} heuristic of \citep{bai2025qwen2}: each frame is isotropically scaled so that the product “\#frames $\times$ pixels” does not exceed $16384{\times}28{\times}28$ (\(\approx\)12.8 M pixels).  
All model weights are fine-tuned with the reinforcement-learning objectives described in Section \ref{method}.  
Optimization uses AdamW (learning-rate $1\!\times\!10^{-5}$, $\beta_1{=}0.9$, $\beta_2{=}0.95$, weight-decay $0.02$) and a cosine decay schedule without warm-up.  
Training is distributed over 4 A100 80 GB GPUs in bfloat16 precision with fully-sharded data-parallelism; gradient accumulation yields an effective batch of 32 video–query pairs.  
A single pass over the training split (\emph{one epoch}) suffices to converge both temporal- and spatial-grounding stages, requiring roughly 280 GPU-hours end-to-end.  
\begin{figure}[t] 
    \centering 
    \includegraphics[width=0.9\textwidth]{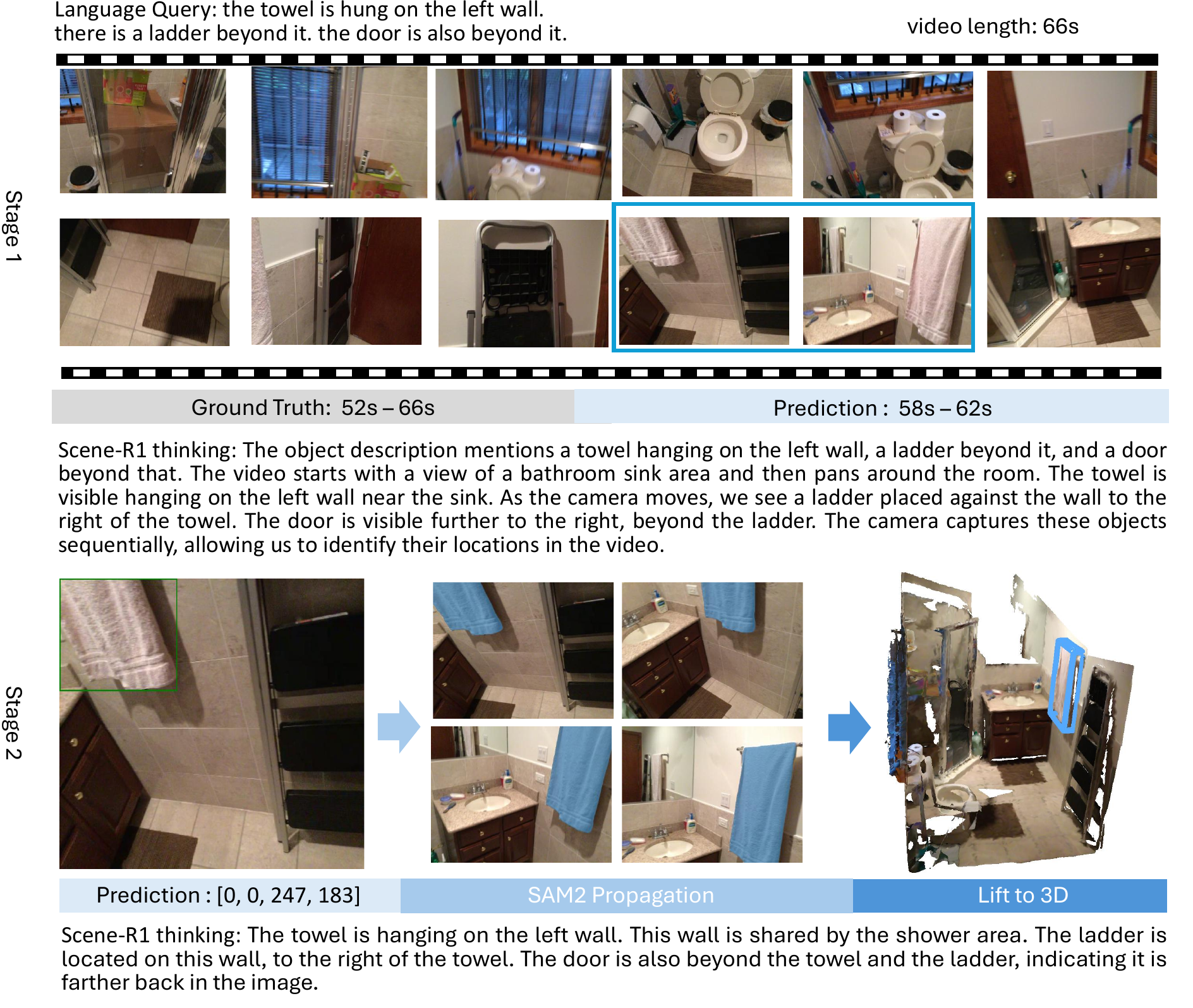} 
    \caption{Visualization of visual grounding.}
    \label{fig:vis_1}
\end{figure}
\begin{table}[t] 
\centering
\caption{Detailed performance comparison on VSI-Bench \citep{yang2025thinking}. Bold values indicate the highest reported score in each column.}
\label{tab:vsibench}
{
\begin{adjustbox}{max width=\linewidth}
\begin{tabular}{l|cccc|cccc|c}
\toprule
\multirow{2}{*}{\textbf{Methods}} & \multicolumn{4}{c|}{\textbf{Numerical Answer}} & \multicolumn{4}{c|}{\textbf{Multiple-Choice Answer}} & \multirow{2}{*}{\textbf{Avg.}} \\
\cmidrule(lr){2-5} \cmidrule(lr){6-9}
& Obj. Count & Abs. Dist. & Obj. Size & Room Size & Rel. Dist. & Rel. Dir. & Route Plan & Appr. Order & \\
\midrule
GPT-4o & 46.2 & 5.3 & 43.8 & 38.2 & 37.0 & 41.3 & 31.5 & 28.5 & 34.0 \\
Gemini-1.5 Pro & 56.2 & 30.9 & 64.1 & 43.6 & \textbf{51.3} & \textbf{46.3} & \textbf{36.0} & 34.6 & \textbf{45.4} \\
LongVA-7B & 38.0 & 16.6 & 38.9 & 22.2 & 33.1 & 43.3 & 25.4 & 15.7 & 29.2 \\
LLaVA-OneVision-7B & 47.7 & 20.2 & 47.4 & 12.3 & 42.5 & 35.2 & 29.4 & 24.4 & 32.4 \\
LLaVA-Video-7B & 48.5 & 14.0 & 47.8 & 24.2 & 43.5 & 42.4 & 34.0 & 30.6 & 35.6 \\
InternVL2-8B & 31.3 & 29.0 & 48.9 & 44.2 & 38.0 & 33.4 & 28.9 & 46.4 & 37.5 \\
LLaVA-OneVision-72B & 43.5 & 23.9 & 57.6 & 37.5 & 42.5 & 39.9 & 32.5 & 44.6 & 40.2 \\
LLaVA-Video-72B & 48.9 & 22.8 & 57.4 & 35.3 & 42.4 & 36.7 & 35.0 & \textbf{48.6} & 40.9 \\
Video-R1 \citep{feng2025video} & 47.0 & 46.1 & 64.2 & 46.5 & 29.1 & 26.7 & 30.9 & 11.7 & 37.8 \\
\textbf{Ours (Scene-R1)} & \textbf{60.8} & \textbf{46.7} & \textbf{66.0} & \textbf{46.9} & 29.0 & 35.9 & 28.9 & 17.0 & 41.4 \\
\bottomrule
\end{tabular}
\end{adjustbox}
}
\end{table}

\subsection{Quantitative Analysis}
\label{sec:main_results}

\subsubsection{3D Visual Grounding}
Table \ref{tab:tab1} summarizes ScanRefer \citep{chen2020scanrefer} grounding accuracy measured as the percentage of predictions whose 3D IoU with the ground-truth box exceeds 0.25 or 0.50.  The upper block of the table lists fully-supervised methods that are trained with dense instance masks and currently define the performance ceiling; the middle block contains systems that forego grounding annotations but still depend on a detector or segmentor pretrained with point-wise 3-D labels to supply proposal boxes and features; the lower block gathers label-efficient or zero-shot approaches. Scene-R1 uses task-level 2D supervision and surpasses VLM-Grounder \citep{xu2025vlm}, by +1.5 and +8.4 percentage points, respectively.  While fully-supervised detectors remain ahead in absolute terms, Scene-R1 demonstrates that a single video-grounded vision–language model, trained end-to-end with lightweight IoU rewards, can deliver competitive 3-D localization without training a dedicated 3D detector, thereby offering a practical route toward scalable 3-D scene understanding.

\subsubsection{Task-driven Affordance Grounding}

Results on SceneFun3D \citep{delitzas2024scenefun3d} are reported in Table \ref{tab:tab3} using mean average precision at IoU thresholds 0.50 and 0.25 ($AP_{50}$ and $AP_{25}$) over points.  The fully supervised upper-bound, OpenMask3D-F, attains 8.0 / 17.5.  The listed OpenMask3D and LERF baselines obtain 0.0 / 0.0 and 4.9 / 11.3, respectively. Scene-R1 obtains 12.0 $AP_{25}$, corresponding to relative gains 6 \% over LERF.  Although a margin remains for the fully supervised detector, these numbers confirm that our video-grounded VLM can localize actionable regions from task descriptions using the proposed video-grounded pipeline.

\begin{table}[htbp]
\centering
\caption{Quantitative results on task-driven affordance grounding.}
\label{tab:tab3}
\begin{adjustbox}{max width=\linewidth}
\begin{tabular}{lcccc}
\toprule
Methods & Supervision & $AP_{50}$ & $AP_{25}$ \\
\midrule
OpenMask3D-F & fully  & 8.0 & 17.5 \\
\hline
OpenMask3D & - & 0.0 & 0.0 \\
LERF & - & 4.9 & 11.3 \\
\textbf{Ours (Scene-R1)} & - & \textbf{6.3} & \textbf{12.0} \\
\bottomrule
\end{tabular}
\end{adjustbox}
\end{table}

\begin{figure}[t]
    \centering
    \includegraphics[width=1\linewidth]{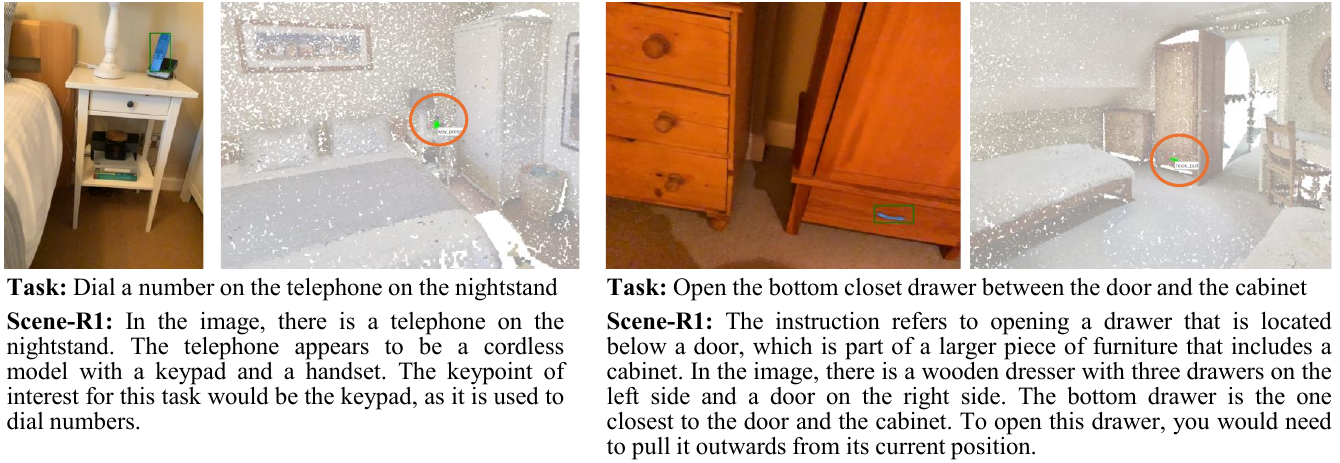}
    \caption{Visualization of task-driven affordance grounding. For each instruction, we show the input RGB view (left) and the reconstructed 3-D scene (right) with the predicted affordance keypoint highlighted (green marker, circled in red). The accompanying natural-language rationales (below each pair) reveal how the model reasons about the constraints.}
    \label{fig:vis_2}
\end{figure}

\subsubsection{3D Question Answering}

Table \ref{tab:tab2} lists results on SQA3D \citep{ma2022sqa3d}, evaluated using exact match accuracy (EM) and its refined version, EM-R, as proposed by LEO \citep{huang2024embodied}.  All baselines are trained with fully supervised geometric signals, whereas Scene-R1 relies only on the weak rewards described in Section \ref{method}.  Under this constraint, Scene-R1 attains an average EM-1 of 49.7\%, outperforming the official SQA3D baseline (46.6\%), and narrowing the gap to more recent detector-based methods, such as LEO \citep{huang2024embodied}. 

We evaluate our model's 3D question answering capabilities on the video-centric benchmark VSI-Bench. As shown in Table~\ref{tab:vsibench}, we compare Scene-R1 against a series of models, where our method achieves a strong average performance of 41.4. Under the weak rewards described in Section~\ref{method}, this excellent performance on a benchmark designed for its native video-stream modality further highlights Scene-R1's effectiveness. The results underscore our model's robust capabilities in handling complex 3D question answering and its emergent visual-spatial intelligence. This finding demonstrates that reinforcement learning with accuracy and format rewards is sufficient to equip a generic vision-language model with competitive 3D reasoning ability, without recourse to point-cloud annotations or task-specific architectural components.

\subsection{Qualitative Analysis}
\label{sec:qualitative}

\subsubsection{3D Visual grounding}
In Fig.~\ref{fig:vis_1} we show a 3DVG example using our method.
\emph{Stage 1} (top) illustrates the temporal grounding stage.  
Scene-R1 selects the interval 58--62\,s (blue bar) from a 66\,s clip. This interval lies within the annotated interval 52--66\,s. These displayed endpoints yield a continuous-time interval IoU of $4/14 \approx 0.286$. This illustrative calculation is distinct from the frame-set metric in Eq.~\eqref{eq:iou_time}, which requires the actual annotated frame indices.  
\emph{Step 2} (bottom) shows the second stage: image grounding.
The predicted 2D box (green) tightly encloses the towel, while the lifted
3D bounding box (blue mask) aligns well with the labelled towel region in the
reconstructed point cloud.  
The generated chain-of-thought explains the spatial hierarchy 
(\emph{“ladder beyond the towel, door further back”}) in natural language,
providing human-readable evidence for the model’s decision.

\begin{table}[t]
\centering
\caption{Ablation study on stage 1: temporal grounding (left) and stage 2: image grounding (right).}
\label{tab:ab2}
\begin{adjustbox}{max width=\linewidth}
\begin{tabular}{l|ccc|cccccc}
\toprule
\multirow{2}{*}{Model} & \multicolumn{3}{c|}{Temporal Grounding} & \multicolumn{2}{c}{ScanRefer} & \multicolumn{2}{c}{SceneFun3D} \\
 & mIoU & R1@0.3 & R1@0.5 & Acc@0.25 & Acc@0.5 & Acc@0.25 & Acc@0.5 \\
\midrule
QWen2.5-VL (zero-shot) & 15.1 & 13.4 & 3.8 & 32.2 & 25.9 & 20.3 & 2.5 \\
QWen2.5-VL (SFT) & 25.4 & 36.3 & 11.0 & 51.7 & 43.1 & 42.9 & 10.9 \\
VLM-Grounder \citep{xu2025vlm} & - & - & - & 62.4 & 53.2  & - & - \\
Ours w/o thinking & 33.1 & 54.8 & 23.4 & 68.6 & 57.3 & 46.5 & 9.1 \\
Ours & \textbf{34.3} & \textbf{56.6}  & \textbf{24.8} & \textbf{70.4} & \textbf{59.4} & \textbf{48.5} & \textbf{11.0} \\
\bottomrule
\end{tabular}
\end{adjustbox}
\end{table}
\begin{figure}[t]
    \centering
    \includegraphics[width=0.9\linewidth]{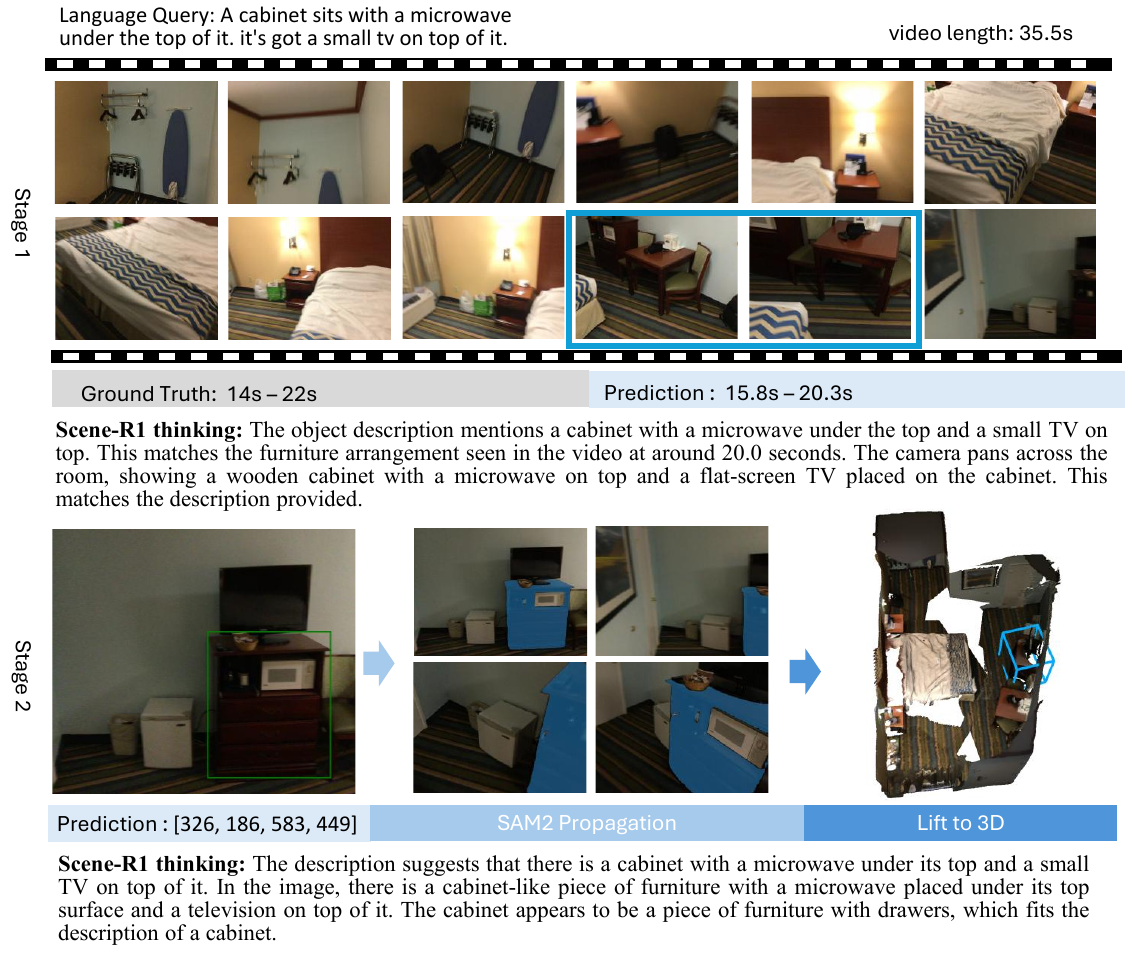}
    \caption{Visualization of 3D visual grounding.}
    \label{fig:supp_1}
\end{figure}

\subsubsection{Affordance grounding}
Figure~\ref{fig:vis_2} visualises two task-driven instructions.
\textit{Dial a number on the telephone on the nightstand}.  
Scene-R1 highlights the telephone keypad—precisely the actionable
sub-region—within both the RGB frame and the 3D cloud, showing that
it can attend to fine-grained keypoints rather than the entire object.
\textit{Open the bottom closet drawer between the door and the cabinet}.  
The model grounds the correct drawer despite multiple similar fronts
and justifies the choice in its CoT by referencing the relative
position between the door and the cabinet.
Across examples, Scene-R1 maintains spatial consistency between its text
rationale, 2D output, and lifted 3D mask—illustrating how the unified
pipeline of Sec.~\ref{method} yields interpretable, cross-modal alignment
through the video-grounded pipeline.

\begin{figure}[t]
    \centering
    \includegraphics[width=0.95\linewidth]{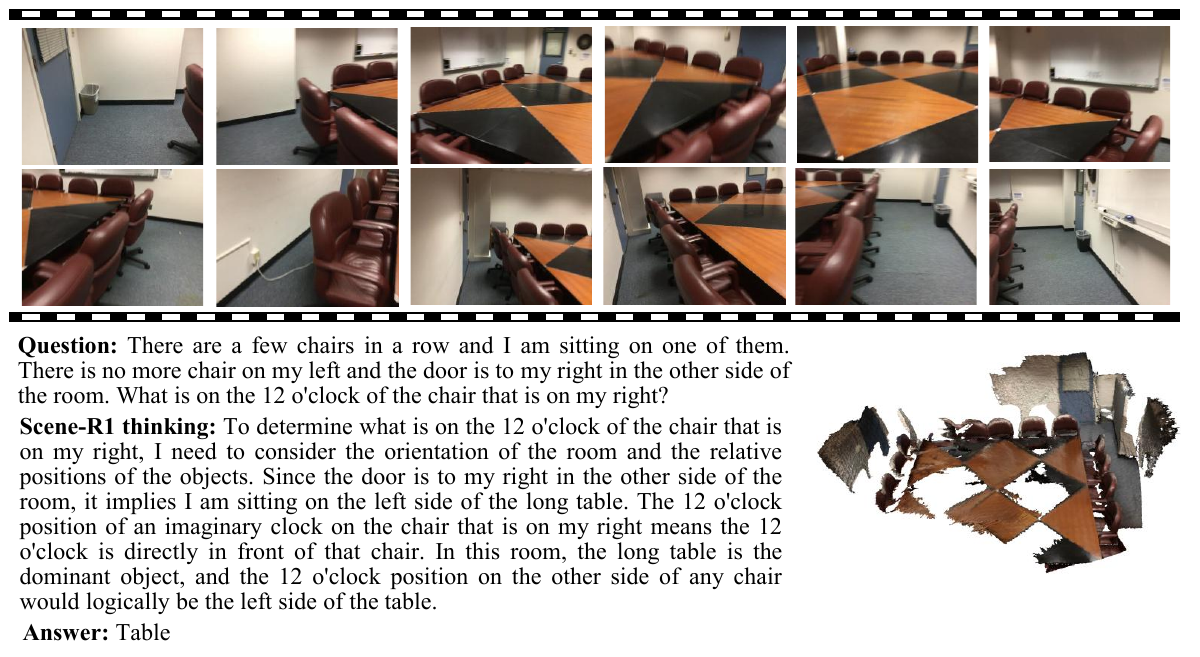}
    \caption{Visualization of 3D visual question answering.}
    \label{fig:supp_2}
\end{figure}
\subsection{Ablation Study}
\label{sec:abl}

Table \ref{tab:ab2} reports improved temporal and image-grounding scores with reinforcement-learning-based optimization. Temporal grounding increases from 15.1 to 34.3 mIoU and from 13.4 to 56.6 R1@0.3 relative to the zero-shot model. The SFT row reports 25.4 mIoU, while removing the thinking prompt yields 33.1 mIoU.
For image grounding, the table reports 70.4 / 59.4\% on ScanRefer and 48.5 / 11.0\% on SceneFun3D at the two stated IoU thresholds. The corresponding SFT values are 51.7 / 43.1\% and 42.9 / 10.9\%. The improvement over SFT is therefore larger on ScanRefer, while the SceneFun3D gain at the higher threshold is 0.1 percentage points. Removing the thinking prompt reduces the reported scores in both stages. These stage-level comparisons complement the end-to-end task results.

\subsection{Visualizations}

Here we present two additional qualitative examples that demonstrate Scene-R1 across different 3D reasoning tasks.
In Fig. \ref{fig:supp_1}, the model receives a relational grounding query and narrows a 35.5\,s video clip to the 15.8--20.3\,s interval containing the relevant scene configuration. It then predicts a tight 2D bounding box around the target cabinet, propagates the mask using SAM2, lifts the result into 3D space, and generates intermediate reasoning traces describing the vertical arrangement of the objects. 
Figure \ref{fig:supp_2} illustrates a spatial reasoning example. Scene-R1 infers from the door position that the camera wearer is seated on the left side of the long table. Based on this orientation, the object located at the 12 o’clock direction relative to the reference chair corresponds to the table. The associated point cloud visualization shows the resulting 3D localization. These examples demonstrate that Scene-R1 can support both 3D grounding and spatial question answering while providing interpretable intermediate reasoning steps.

\section{Conclusion}

We introduced Scene-R1, a video-grounded framework for 3D scene understanding that uses task-level 2D and language supervision without training a dedicated 3D instance detector. Our method decomposes the task into a temporal grounding stage and a spatial refinement stage, both optimized using group-relative policy optimization with lightweight rewards based on \textit{IoU} and output format consistency. Unlike previous 3D vision-language pipelines, Scene-R1 does not require proposal networks or pre-trained 3D instance segmentation models. Instead, it learns directly from RGB-D video while producing intermediate reasoning traces during inference. Experimental results show that reinforcement-learning-based training on video data can support effective and annotation-efficient 3D scene understanding.

\section{Limitations}
The reported experiments concern indoor datasets. Extending the method to outdoor or dynamic scenes remains future work. Reconstruction depends on depth measurements, camera calibration, segmentation, and temporal selection; errors in these components can propagate to the final localization. Generated reasoning traces support inspection of intermediate predictions, but their presence alone does not establish causal faithfulness.

\section*{Acknowledgments}
This work was supported in part by the Guangdong S\&T Programme (Grant No. 2024B0101030002), the Basic Research Project of Hetao Shenzhen-HK ST Cooperation Zone (No. HZQBKCZYZ-2021067), the Shenzhen Outstanding Talents Training Fund (No. 202002), and the NSFC (Grant Nos. 62293482, 62573371, 61931024, and 12326610). This work was also supported by the Guangdong Province Radio Science Data Center (Grant No. 2025B1212070001), the Shenzhen General Program (No. JCYJ20220530143600001), the Guangdong Research Projects (Nos. 2017ZT07X152 and 2019CX01X104), the Guangdong Provincial Key Laboratory of Future Networks of Intelligence (Grant No. 2022B1212010001), the Shenzhen Key Laboratory of Big Data and Artificial Intelligence (Grant No. SYSPG20241211173853027), the National Key Research and Development Program of China (Nos. 2025YFF0515300 and 2025YFF0515304), the Open Project Program of Key Laboratory of Tibetan Information Processing, Ministry of Education (Grant No. QHSFCS2606), the Shenzhen-Hong Kong Joint Funding (No. SGDX20211123112401002), and the Tencent \& Huawei Open Fund (Nos. 2024E0009 and 202301030019).

\bibliographystyle{iclr2026_conference}
\bibliography{scene_r1_references}

\clearpage
\appendix
\section{Inference Interface}
The accompanying code exposes temporal grounding, image grounding, question answering, RGB-D lifting, and SAI3D candidate matching. Model checkpoints, selected frames, and propagated target masks are supplied separately. The following prompts specify the accompanying inference interface; \texttt{[QUERY]} is replaced by the input query.

\subsubsection*{Temporal grounding}
\begin{tcolorbox}[]
Locate the object described as "[QUERY]" in the video. Output reasoning in \texttt{<think></think>} and the time interval in seconds as \texttt{<answer>start to end</answer>}.
\end{tcolorbox}

\subsubsection*{Image grounding}
\begin{tcolorbox}[]
Outline the object described as "[QUERY]" in the image. Output reasoning in \texttt{<think></think>} and exactly one JSON object in \texttt{<answer></answer>}: \texttt{\{"bbox\_2d": [x1, y1, x2, y2], "label": "object"\}}. Use pixel coordinates of the image you see, not 0--1 or 0--1000 normalized coordinates.
\end{tcolorbox}

\subsubsection*{Question answering}
\begin{tcolorbox}[]
[QUERY] Output reasoning in \texttt{<think></think>} and the short final answer in \texttt{<answer></answer>}.
\end{tcolorbox}

\section{Use of Large Language Models}
Large language models assisted with language editing, manuscript formatting, and organization of the accompanying code. This assistance does not independently validate the experimental results; the authors are responsible for the final manuscript and implementation.

\end{document}